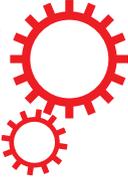

# Clinical Assistant Diagnosis for Electronic Medical Record Based on Convolutional Neural Network

Zhongliang Yang[1,2], Yongfeng Huang[1,2], Yiran Jiang[3], Yuxi Sun[3], Yu-Jin Zhang[1] & Pengcheng Luo[4]

Automatically extracting useful information from electronic medical records along with conducting disease diagnoses is a promising task for both clinical decision support(CDS) and neural language processing(NLP). Most of the existing systems are based on artificially constructed knowledge bases, and then auxiliary diagnosis is done by rule matching. In this study, we present a clinical intelligent decision approach based on Convolutional Neural Networks(CNN), which can automatically extract high-level semantic information of electronic medical records and then perform automatic diagnosis without artificial construction of rules or knowledge bases. We use collected 18,590 copies of the real-world clinical electronic medical records to train and test the proposed model. Experimental results show that the proposed model can achieve 98.67% accuracy and 96.02% recall, which strongly supports that using convolutional neural network to automatically learn high-level semantic features of electronic medical records and then conduct assist diagnosis is feasible and effective.

Automatically extracting useful information from electronic medical records (EMRs) with implementing apropriate diagnosis is one ultimate goal of intelligent medical construction[1]. It is such a meaningful and promising task that it can not only effectively improve working efficiency but also reduce misdiagnosis rate of doctors for making a diagnosis[2–4]. At the same time it can help us to better understand the clinical manifestations of various diseases[5–8], and even find the relation between the various diseases[9,10]. Previous works show that such models can even sometimes outperforming experienced doctors[11] in improving teaching practice[12,13] and assisting diagnosis[14–17].

However, it is extremely challenging for the following reasons: firstly, medical records contain complex reports of patients which include patient's statements, vital signs, history of treatment, history of allergies and so on. The varieties of which make it arduous to filter and represent complicated information. Secondly, it is possible that different patients with the same diseases may have disparate performances or symptoms, which makes it difficult to establish a unified rule to predict the possibility of incidence and analyze information of diseases. Thirdly, notes and reports in EMRs are likely to be written by different doctors from diverse departments. Thus, even if some words or records seem to have the same meaning, there would still exist dissimilar writing expressions due to different writing habits, which increases the difficulty of identifying the actual symtoms as well as reaching an accurate conclusion.

Most of existing platforms are rule-based methods, which can also be called expert systems. Expert systems are designed to solve complex problems by reasoning through bodies of knowledge, represented mainly as hand-crafted rules. They carry out rule matching on each input electronic medical record in order to chase down the disease which fits these diagnosis rules best and make a diagnose for the disease. This kind of methods have made great achievements in the field of medical auxiliary diagnosis[18–20]. These models hope to imitate the logical reasoning process in the diagnosis procedure of a doctor which makes its diagnosis more logical. However, it weakens the effects in viable utilization owing to some deficiencies of the model itself. Firstly, it cannot solve the problem like semantic ambiguity. Hence, it should contain all potential descriptions as far as possible while setting up the knowledge base, which may cause redundancies in the knowledge base and result in low efficiencies. For example, it was estimated that about fifteen person-years were spent building the Internist-1/QMR knowledge base for internal medicine[21]. It may also lead to decline in matching accuracy because of not considering all possible conditions. Secondly, a variety of hospitals and departments encounter a wide range of cases which may have great disparities. It is fairly complicated and adverse to manage and maintain the knowledge base constructed by hundreds and thousands of diseases from such various departments, which leads to low efficiencies.

[1]Department of Electronic Engineering, Tsinghua University, Beijing, 100084, China. [2]Tsinghua National Laboratory of Information Science and Technology, Beijing, 100084, China. [3]International School, Beijing University of Posts and Telecommunications, Beijing, 100876, China. [4]Huangshi Central Hospital of Edong Healthcare Group, Hubei Polytechnic University, Hubei, 435000, China. Correspondence and requests for materials should be addressed to Y.H. (email: yfhuang@tsinghua.edu.cn)

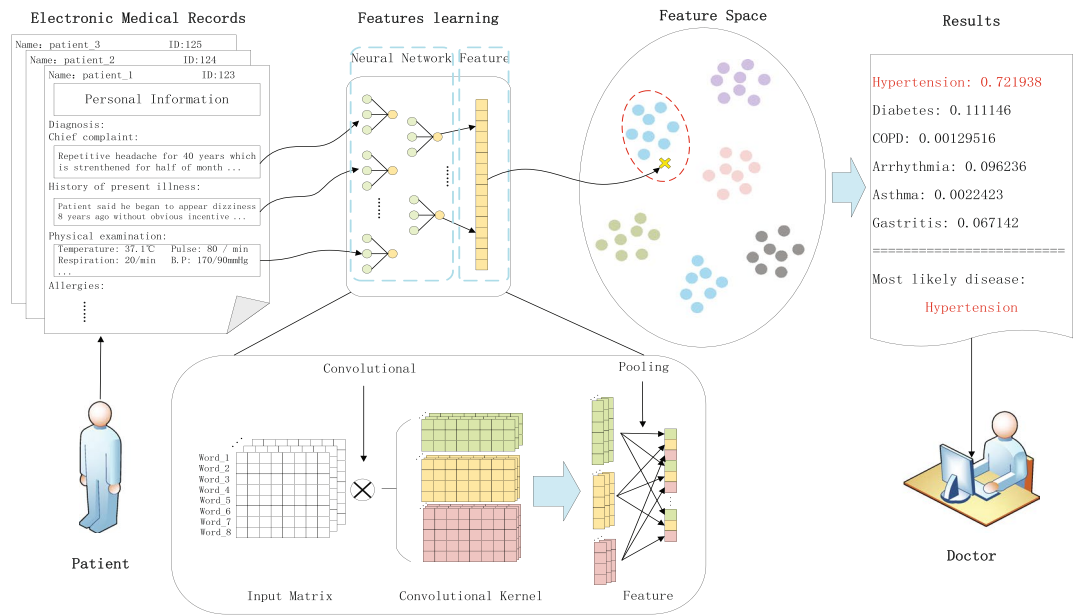

**Figure 1.** The overall framework of the proposed model. We use the convolutional neural network to extract the semantic feature vectors of unstructured electronic medical records and map them to the feature space, finally we use the classifier to calculate the probable probability of each disease and select the highest probability of the disease as the auxiliary diagnosis of our model.

High level semantic understanding for medical record texts has always been hard because of its high coding degree[22,23]. In recent years, with the development of natural language processing, there has been an increasing number of auxiliary diagnostic methods based on semantic analysis algorithm[16,22,24]. These kinds of methods try to conduct a high level semantic understanding on EMRs, which mainly draw on natural language processing related technology[22]. They hope to help the computer better understand the semantics of electronic medical records, and then make a diagnosis accordingly. The ultimate goal they hope to achieve is far from easy to achieve. During these years, with the extensive adoption of deep neural network technology in the field of natural language processing, the application of deep neural network on semantic understanding with analyzing texts has become a popular research[23,25,26].

To achieve the ultimate goal, in this study, we applied a multi-layer convolutional neural network for high level semantic understanding for electronic medical records, which can then be used for disease diagnoses. In the past few years, convolutional neural network has made notable progress in fields such as computer vision[27–29] and natural language processing[23,26]. The incremental advancement of CNN is likely to benefit the development of new technology and inventions in other fields. A large number of researches and applications have shown that the convolutional neural network has a powerful ability in feature extractions and expressions[27,30], which does not require hand-designed features but carries out self-learning through plenty of data. Previous studies have shown that neural network can represent the words in the texts into a dense vector through learning and mapping them into a continuous vector space[31–35]. In this vector space, semantically similar words are distributed in the same region[33]. Thus, even if the two sections of the text are not the same, as long as the expressions are of the same meaning, they will have similar mathematical expressions, reflected in the semantic space very close[32]. This can greatly alleviate the problem of semantic ambiguity, and is more efficient than the model based on knowledge base. So we don't need to build a large number of complex rules or knowledge base to guide how the model decides, but the model itself can automatically extract useful information from the electronic medical records by self-learning, and then conduct disease diagnoses based on these information. This makes our model lighter and more efficient than the knowledge base-based model. The overall framework of our model is shown in Fig. 1. The input of our model is an electronic medical record and the output is the probability of diseases we predicted.

## Results

**Data Preparing.** To promote the development of the related fields, in this study, we collected and released a large real-world electronic medical records dataset (C-EMRs) collected from Huangshi Central Hospital in China. It has a total number of 18,590 EMRs and contains the most common diseases of each department, which are Hypertension, Diabetes, Chronic Obstructive Pulmonary Disease (COPD), Gout, Arrhythmia, Asthma, Gastritis, Stomach Polyps. After expunging personal information, each electronic medical record includes thirteen items: chief complaint, physical examination, history of present illness and so forth. Each electronic medical

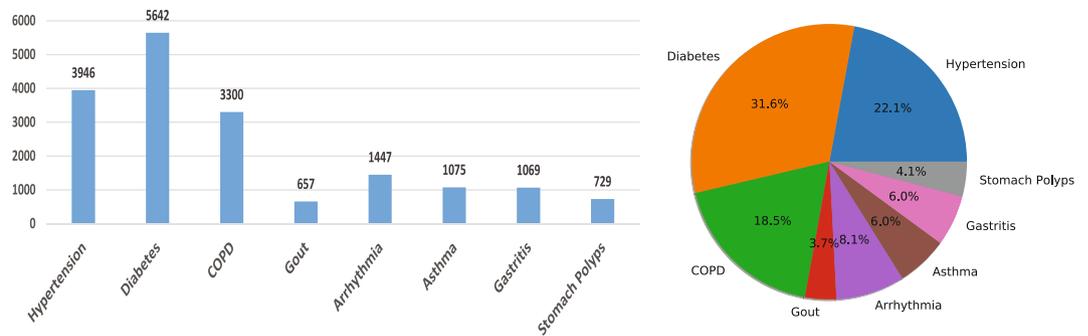

**Figure 2.** The number and proportion of each disease in C-EMRs.

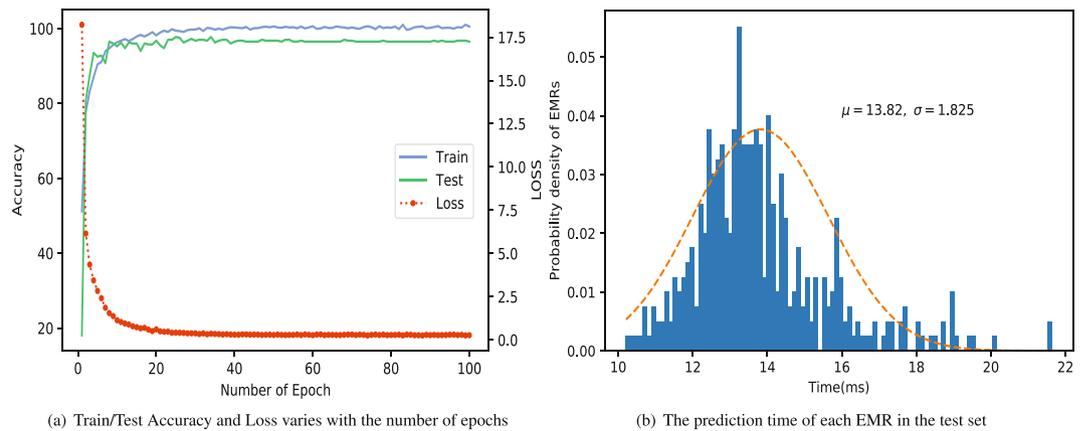

(a) Train/Test Accuracy and Loss varies with the number of epochs

(b) The prediction time of each EMR in the test set

**Figure 3.** The processing of training. (**a**) Shows the accuracy of train/test set and the loss of the training set varies with the number of epochs. (**b**) Shows the prediction time of each EMR in the test set.

|  | Hypertension | Diabetes | COPD | Arrhythmia | Asthma | Gastritis | Total |
|---|---|---|---|---|---|---|---|
| Trianing set | 1250 | 1350 | 1250 | 1200 | 1000 | 950 | 7000 |
| Test set | 68 | 68 | 68 | 62 | 69 | 65 | 400 |

**Table 1.** The number of electronic medical records of each disease in the training set and test set, and the percentage of test data relative to training data of each disease.

record corresponds to a result of doctor's diagnosis, which will be used as the label for each EMR samples during the training process. Due to the possibility of a patient with multiple diseases, it is possible that the two electronic medical records have the same content, but the diagnostic results are different. In our dataset, there are altogether 447 patients whose situations are consistent with what is mentioned above. The number and proportion of each disease are shown in Fig. 2.

The electronic medical records number of different disease in C-EMRs are imbalance. For diabetes, there are 5642 medical records, but for gout there are only 657. In order to avoid biases and to ensure that there is enough training data, we choose the diseases that has more than 1,000 records to form the training set, which are hypertension, diabetes, COPD, arrhythmia, asthma and gastritis. Also in order to prevent that the training dataset has too much biases, we randomly selected almost the same number records of these diseases as training and testing data. So finally the training data for our model are 7000 EMRs, and another 400 EMRs for testing, which are distributed as Table 1.

**Experiment Results.** We use stochastic gradient descent with momentum 0.9 to train parameters of our network. Our model can quickly converge during the training processing, training after about 20 epochs (one epoch means that all the training samples finish one training session) can reach a steady state with high accuracy and the loss curve is very smooth, which can be seen in Fig. 3(a). From Fig. 3(b) we can see that the prediction time of each electronic medical record is mainly between 10 and 20 milliseconds, which can be predicted in real time.

In Table 2, the Precision, Recall, F1-score and Accuracy of four machine learning algorithms, which are Support Vector Machine (SVM), Multinomial Naïve Bayes(MultinomialNB), Logistic Regression and k-NearestNeighbor, as well as our proposed model are reported. These four machine learning algorithms that we compared to have been applied to the auxiliary diagnosis of electronic medical records in some previous related works and have achieved good results[36–38]. From the results shown in Table 2 we can see that our model has achieved the best effect on each evaluation method. On the test set, our model achieves a 98.67% accuracy and a recall rate of 96.02%, which strongly proves that CNN do have stronger capability of information extraction from texts than other algorithms. Table 3 shows the average prediction time of different methods for each EMR in test set. From Table 3 we can see, the average diagnostic time of our model for each electronic medical record is only 13.82 milliseconds, which indicates that our model can be very efficient in the diagnosis process.

| Method | Training set | | | | Testing set | | | |
|---|---|---|---|---|---|---|---|---|
| | Precision | Recall | F1-score | Accuracy | Precision | Recall | F1-score | Accuracy |
| SVM | 0.96 | 0.95 | 0.96 | 0.9549 | 0.93 | 0.93 | 0.93 | 0.9315 |
| MultinomialNB | 0.93 | 0.92 | 0.92 | 0.9236 | 0.87 | 0.86 | 0.86 | 0.8600 |
| LogisticRegression | 0.93 | 0.93 | 0.93 | 0.9293 | 0.92 | 0.92 | 0.92 | 0.9175 |
| KNeighborsClassifier | 0.89 | 0.89 | 0.89 | 0.8911 | 0.90 | 0.89 | 0.89 | 0.8925 |
| CNN | **0.9947** | **0.9946** | **0.9946** | **0.9982** | **0.9594** | **0.9602** | **0.9596** | **0.9867** |

**Table 2.** Results of different methods, where "CNN" indicates the performance of the proposed model.

| Method | SVM | MultinomialNB | LogisticRegression | KNeighborsClassifier | CNN |
|---|---|---|---|---|---|
| Time(ms) | $180.5 \pm 2.92$ | $172.5 \pm 2.55$ | $167.5 \pm 1.58$ | $205.0 \pm 1.0$ | **$13.82 \pm 1.83$** |

**Table 3.** The average prediction time of different methods for each EMR in test set, where "CNN" indicates the performance of the proposed model.

As we have mentioned before, our model can automatically extract high-level semantic features from electronic medical records and map them to a high-dimensional feature space (usually hundreds to thousands of dimensions). We can use t-Distributed Stochastic Neighbor Embedding (t-SNE)[39,40] technique for the dimensionality reduction and visualization of these high-dimensional feature vectors, which can be find in Fig. 4. In this feature space, each point represents an electronic medical record and different colors indicate different diseases. At the beginning of training (Epoch = 0), since the model parameters are randomly initialized, all the electronic medical records in the feature space are randomly distributed and indivisible. After 5 epoch, electronic medical records of different diseases began to have a trend of separation. After 10 epoch, the electronic medical records of all kinds of diseases have been separated, except for some areas and the edge of each category. When the training reaches 100 epoch, we can clearly see that the samples of each disease have been completely separated, and the electronic medical records of the same disease are also gathered together. After training, electronic medical records belong to the same kind of disease distribution in the same area. Considering that some patients may also suffer from a variety of diseases, there will be individual records mixed with other categories. For each inputted electronic medical record, we mapped it to the feature space, and by analyzing its location distribution in the feature space, we can calculate the possibility of which disease it belongs to.

## Discussion

Automatic extraction of useful information in electronic medical records is of great significance and value for the study of clinical treatment and related diseases[1,5–7]. The current clinical diagnosis model or system is mostly based on the large-scale medical knowledge base of human construction[18,41,42]. Through the association extraction of electronic medical records and the rule matching with the knowledge base, the electronic medical records are analyzed and the clinical auxiliary diagnosis is provided. This kind of method is usually of heavy workload[21] and the actual effect is not very satisfactory. In this study, we propose a method of information extraction and analysis of electronic medical records using convolutional neural network, and finally conduct clinical auxiliary diagnosis. Comparing with other machine learning algorithms, our model is proved superior to other algorithms on various metrics (Table 2). The high precision (95.94%) achieved by our model means that the probability of misdiagnosis of our model is very low, which is extremely important for practical use. At the same time, the recall (96.02%) of our model is also high, which means that the probability of missed diagnosis is extremely low in our model. Combined with these test results, we can find that our model has significantly practical value.

It is worth noting that our model does not require human building large scale knowledge bases and complex rules, since all the model parameters and features are automatically learned from a large number of historical electronic medical records, which makes our model quite lightweight and fairly practical. At the same time, our model is very efficient, through testing we found that the average prediction time of each electronic medical record is 13.82 milliseconds, which outperforms other machine learning methods (SVM: 180.5 ms, MultinomialNB: 172.5 ms, LogisticRegression: 167.5 ms, KNeighborsClassifier: 205.0 ms), which has been shown in Table 3.

These results strongly support that it is feasible and effective to use the convolutional neural network to automatically learn high-level semantic features of electronic medical records and then conduct assist diagnosis. Based on these advantages, our model can effectively improve the clinical diagnostic efficiency of doctors. At the same time, because our model is affirmed by a large number of historical diagnostic medical records, it can effectively reduce the possibility of misdiagnosis.

As the results shown in Fig. 3, our model can effectively extract high-level semantic features of electronic medical records and map them into high-dimensional feature space. In this feature space, electronic medical records of different diseases have different distribution, and the electronic medical records of the same disease are gathered together. By analyzing this feature space (we can also call it "disease space"), we may even be able to help clinicians better understand the relation between various diseases and what tendencies are likely to occur in the same disease, which would be one of the most promising aspects of the proposed model. We hope that our research will not only help clinicians better make clinical diagnosis, on the other hand, we hope to help clinicians further understand the various clinical diseases from another perspective.

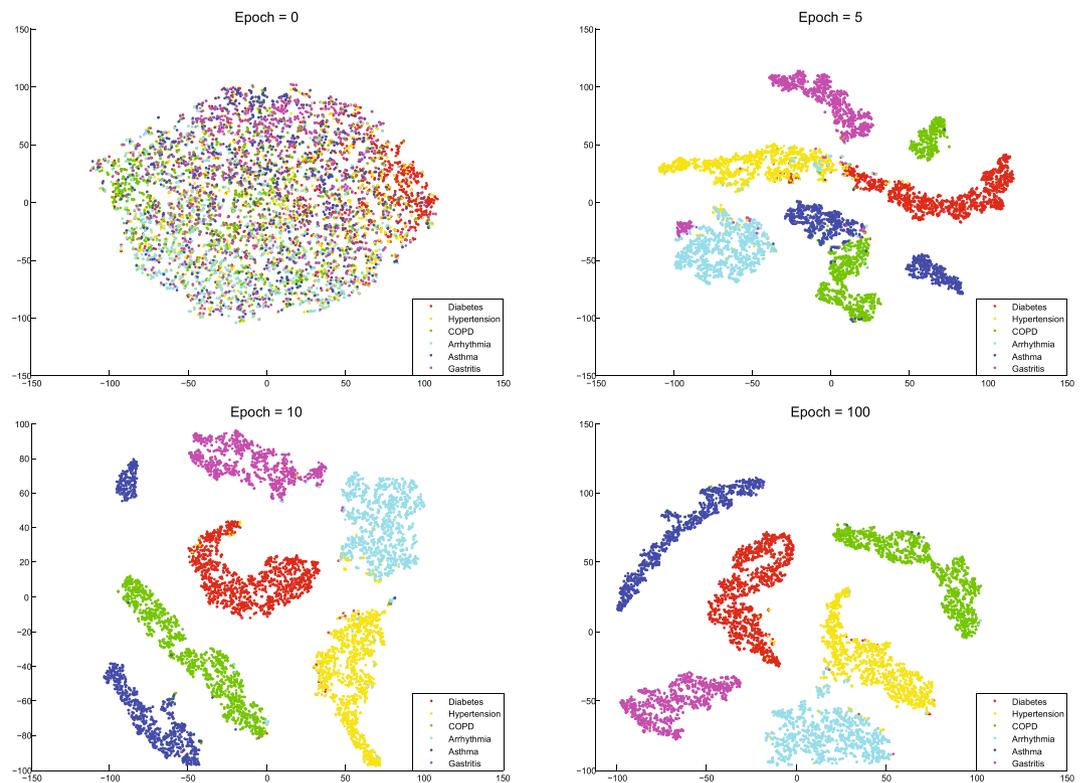

**Figure 4.** The change of feature space with the training process. In this feature space, each point represents an electronic medical record and different colors indicate different diseases. At the beginning of training (Epoch = 0), since the model parameters are randomly initialized, all the electronic medical records in the feature space are randomly distributed and indivisible. After 5 epoch, electronic medical records of different diseases began to have a trend of separation. After 10 epoch, the electronic medical records of all kinds of diseases have been separated, except for some areas and the edge of each category. When the training reaches 100 epoch, we can clearly see that the samples of each disease have been completely separated, and the electronic medical records of the same disease are also gathered together.

Although we have made gratifying achievements, we still should consider some limitations on the current exploratory reseach. Firstly, the sample types used for model training and testing are not enough as we used only several most common and medically different diseases. Therefore, in the future research, we'll try to do more research, including more types of diseases, and more similar diseases, such as diabetes I and II. Secondly, in this study, we only consider three main contents of electronic medical records: chief complaint, history of present illnesses and physical examination. Although these three items are likely the most crucial ones, other contents corresponding to this record are equally important. In the future research, we will take into account comprehensive contents in the electronic medical record, even including other diagnostic information, such as medical images.

In summary, the major contributions of this paper are as follows. Firstly, we designed and implemented an auxiliary diagnosis model for electronic medical records based on convolution neural network. We hope this model can not only effectively improve working efficiency but also reduce misdiagnosis rate of doctors for making a diagnosis. Since our model can conduct high level semantic understanding of the electronic medical records, we also hope that it can help doctors to better understand the clinical manifestations of various diseases, and even the relation between the various diseases. Secondly, in order to promote the development of the related fields, we collected and released a large real-world electronic medical records dataset (C-EMRs). It has a total number of 18,590 EMRs and contains the most common diseases of each department, which are Hypertension, Diabetes, Chronic Obstructive Pulmonary Disease (COPD), Gout, Arrhythmia, Asthma, Gastritis, Stomach Polyps. Thirdly, we tested and evaluated the proposed auxiliary diagnosis model for electronic medical records based on CNN on this dataset. The test results show that the method has high diagnostic efficiency (13.82 milliseconds costs for each EMR prediction) and diagnostic accuracy (acc: 98.67%, recall: 96.02%). Although our model still has space for further improvement, it has shown significant and practical value for clinical research. We hope that our work will serve as a guide for future related work and help promote the further development of the auxiliary diagnosis of electronic medical records.

## Method

**Model Structure and Analysis.** In this study, we propose a method using convolutional neural network to extract features from electronic medical records and conduct disease prediction. The input of the proposed model is an electronic medical record and the output is the prediction probability of diseases. The final structure of the convolutional neural network used in this study is as follows: an embedding layer, a convolutional layer with three different sizes of convolutional kernels, an average pooling layer and a fully connected layer following with a softmax classification. The embedding layer transforms the inputted EMR text into a two-dimensional matrix form which is suitable for the processing of convolution. The convolutional layer is used to extract features from the input matrix and convolution kernels of different sizes can learn different context related features. The pooling layer is served for down sampling the features, which can enhance the robustness of the model and significantly influence the performance[27,43]. The purpose of the fully connected layer is to fuse all these features and pass them to the softmax classifier for disease prediction. The softmax classifier, whose parameters have been learned during the training process, calculates the correlation between the input feature vector and the various diseases, and finally concludes the probability value of each disease. The practical parameters setting will be given in the next section "Experiment Setting".

For each of the structured medical records inputted, we first make it unstructured by connecting each of its contents to form a whole passage. For each passage $S$, we illustrate it with a matrix $X \in \mathbb{R}^{N \times D}$, as shown in Equation (1), where the $i$-th row indicates the $i$-th word in passage $S$, each word is represented as a $D$-dimension vertor which is randomly initialized, that is

$$X = \begin{bmatrix} Word_1 \\ Word_2 \\ \vdots \\ Word_N \end{bmatrix} = \begin{bmatrix} x_{1,1} & x_{1,2} & \cdots & x_{1,D} \\ x_{2,1} & x_{2,2} & \cdots & x_{2,D} \\ \vdots & \vdots & \ddots & \vdots \\ x_{N,1} & x_{N,2} & \cdots & x_{N,D} \end{bmatrix} \tag{1}$$

Generally, let $X_{i:j}$ refer to the matrix which consists of the words vectors from the $i$-th word to the $j$-th word, that is:

$$X_{i:j} = \begin{bmatrix} Word_i \\ Word_{i+1} \\ \vdots \\ Word_j \end{bmatrix} = \begin{bmatrix} x_{i,1} & x_{i,2} & \cdots & x_{i,D} \\ x_{i+1,1} & x_{i+1,2} & \cdots & x_{i+1,D} \\ \vdots & \vdots & \ddots & \vdots \\ x_{j,1} & x_{j,2} & \cdots & x_{j,D} \end{bmatrix} \tag{2}$$

The convolution layer contains convolution kernels of multiple sizes, and each size contains multiple number of convolution kernels. The width of each convolution kernel is the same as the width of the input matrix. Suppose that the height of the $k$-th convolution kernel is $H$, the convolutional kernel can be expressed as $W^k \in \mathbb{R}^{H \times D}$, that is

$$W^k = \begin{bmatrix} w^k_{1,1} & w^k_{1,2} & \cdots & w^k_{1,D} \\ w^k_{2,1} & w^k_{2,2} & \cdots & w^k_{2,D} \\ \vdots & \vdots & \ddots & \vdots \\ w^k_{H,1} & w^k_{H,2} & \cdots & w^k_{H,D} \end{bmatrix} \tag{3}$$

Convolution operation is a feature extraction process for the elements in the local region of the input matrix. For example, when $w^k_{1,1}$ and $x_{1,1}$ coincide, then the feature $c^k_1$ extracted from $X_{1:H}$ by the convolutional kernel can be:

$$c^k_1 = f\left(\sum_{i=1}^{H}\sum_{j=1}^{D} w^k_{i,j} \cdot x_{i,j} + b^k_{i,j}\right), \tag{4}$$

where the weight $w^k_{i,j}$ denotes the importance of the $j$-th value in the $i$-th word vector, $b^k_{i,j}$ is the bias term and $f$ is a nonlinear function, here we follow previous works[27] and use ReLu function as our nonlinear function, which is defined as

$$y = ReLu(x) = max(0, x). \tag{5}$$

The convolution process is that the convolution kernel $W^k$ slides from the top to the bottom on the input matrix $X$ with a certain step $T_c$, and calculates the features of each local region. Finally, the document feature extracted by convolution kernel $W^k$ is:

$$C^k = \left[c^k_1, c^k_2, \ldots, c^k_{\frac{N-H+1}{T_c}}\right]^T. \tag{6}$$

The pooling layer can reduce the number of neural network parameters while maintaining the overall distribution of the data, which can effectively prevent the model over-fitting and improve the robustness of the model[27,43]. The pooling operation is very similar to the convolution operation, the only difference is that it only calculates the average or maximum value of the local area. We conduct a max pooling operation after each convolution operation on the feature $C^k$, suppose the height of a pooling kernel is $H_p$ and the step size is $T_p$, then the output is:

$$M^k = [m_1^k, m_2^k, \ldots, m_{N_p}^k]^T, \quad (7)$$

where

$$m_i^k = max(c_i^k, c_{i+1}^k, \ldots, c_{i+H_p-1}^k). \quad (8)$$

$$N_p = \frac{\frac{N-H+1}{T_c} - H_p + 1}{T_p}, \quad (9)$$

The process described above is a process in which one convolution kernel $W^k$ produces one feature $M^k$. After all the convolution and pooling operations have been completed, all the extracted features are concatenated end to end to obtain the feature vector of the entire EMR, which can be indicated as

$$F^T = [F_1^T; F_2^T; \ldots; F_l^T], \quad (10)$$

where $F_i = M^i$, $l$ indicates the number of the features.

Fully connected layer is used to further blend features and extract higher-level features. By defining a weight matrix $W_F$, we compute the weighted sum of each feature element and obtain the final feature representation of the inputted text $S$:

$$y = W_F \cdot F + b_f, \quad (11)$$

where $W_F$ and $b_f$ are learned weight matrix and bias, the values in weight matrix $W_F$ reflect the importance of each feature. The dimension of the output vector $y$ is $L$, which is the number of labels. In our realization, $L$ is the number of diseases required to be predicted. We then pass the vector $y$ through the softmax classifier to get the predicted probability of each disease:

$$P_i = \frac{\exp(y_i)}{\sum_{j=1}^{L} \exp(y_j)}, \quad (12)$$

where $P_i$ indicates the prevalence of the $i$-th disease corresponding to the input medical records.

In the process of training, we update network parameters through applying backpropagation algorithm, and the loss function of the whole network consists of two parts, one is the error term and the other is the regularization term, which can be described as:

$$LOSS = \sum_{num} (\mathbf{P} - \mathbf{T})^T \cdot (\mathbf{P} - \mathbf{T}) + \|W_F\|_2 = \sum_{num} \sum_{i=0}^{L} (P_i - T_i)^2 + \|W_F\|_2, \quad (13)$$

where $num$ is the batch size of EMRs, P indicates the output of the classifier, and each element as $P_i$ represents the prevalence of the $i$-th disease. T is the target value which corresponds to the doctor's diagnosis result. For instance, if this medical records corresponds to the $t$-th disease, the value of the $t$-th element in vector $T$ is 1, and the remaining values are 0. The error term in the loss function calculates the mean square error (MSE) between the prediction vector and the actual label. We hope that through the self-learning of the model, the mean square error gets smaller and smaller, that is, the prediction results are getting closer to the real values. In order to strengthen the regularization and prevent overfitting, we adopted the dropout mechanism and a constraint on l2-norms of the weight vectors during the training process. Dropout mechanism means that in the training process of deep learning network, the neural network unit is temporarily discarded from the network, i.e. set to zero, according to a certain probability. This mechanism has been proved to effectively prevent neural network overfitting, and significantly improve the model's performance[27,44,45].

We train our model by minimizing the *LOSS* function over a batch size number of samples. We use stochastic gradient descent with momentum 0.9 to train the parameters of our network. The update rule for weight $w$ is:

$$w_{i+1} = w_i + \alpha \cdot V_i - \lambda \cdot \left\langle \frac{\partial L}{\partial w_i} \right\rangle \bigg|_{D_i}, \quad (14)$$

where $i$ is the iteration idex, $\alpha \in (0, 1]$ is the momentum factor, $V$ is the momentum variable, $\lambda$ is the learning rate, and $\left\langle \frac{\partial L}{\partial w_i} \right\rangle \bigg|_{D_i}$ is the average over the $i$-th batch $D_i$ of the derivative of the *LOSS* function with respect to w, evaluated at $w_i$.

**Experiment Setting.** After removing the patient's private information from EMRs, combined with the doctor's advice, we mainly used chief complaint, history of present illnesses and physical examination in EMRs as our input text by just connecting each of its contents to form a whole passage. Then we made use of the most commonly used Chinese text segmentation tool, which is Jieba[46], to do the word segmentation on the input texts. After that, we built the dictionary based on the dataset and counted the length of each passage. We only consider the words that appear more than five times and the others will be remarked as a character "⟨unk⟩". So finally we get 17,274 unique words in our dictionary.

Since our model requires the input matrix be of a certain size, that is, the length of the input text should be constant. We design multiple sets of comparision experiments to choose the best value of this super-parameters. According to the experiment results, we finally fix each input electronic medical record text into 130 words. Less than 130 words will be padding with zero, and more than 130 words will be discarded.

For the input of our model, we map each word to a vector of 300 dimensions which are randomly initialized, so the dimension of input matrix will be $130 \times 300$. The width of the convolutional kernel is the same as the input matrix, thus 300. However, the height of the convolutional kernel is not fixed, we set the kernel heights to be 4, 5, 6 by comparing the results of different kernel sizes, and each of the different heights has 128 convolution kernels. The dimension of the feature extracted for each EMR is $3 \times 128 = 384$ and the dimension of the output vector is six, corresponding to six diseases that require diagnosis, so the weight matrix $W_F$ of fully connected layer would be $W^F \in \mathbb{R}^{384 \times 6}$.

**Evaluation Method.** We use several evaluation indicators commonly used in classification tasks to evaluate the performance of our model, which are precision, recall, F1-score and accuracy. Their conceptions and formulas are described as follows:

- Precision measures the proportion of positive samples in the classified samples.

$$Precision = \frac{TP}{TP + FP}. \tag{15}$$

- Recall measures the proportion of positives that are correctly identified as such.

$$Recall = \frac{TP}{TP + FN}. \tag{16}$$

- F1-score is a measure of a test's accuracy. It considers both the precision and the recall of the test. The F1 score is the harmonic average of the precision and recall, where an F1 score reaches its best value at 1 and worst at 0.

$$F1 - score = \frac{2 \times Precision \times Recall}{Precision + Recall}. \tag{17}$$

- Accu3racy measures the proportion of true results (both true positives and true negatives) among the total number of cases examined

$$Accuracy = \frac{TP + TN}{TP + FN + FP + TN}. \tag{18}$$

where TP (True Positive) represents the number of positive samples that are predicted to be positive by the model, FP (False Positive) indicates the number of negative samples predicted to be positive, FN (False Negative) illustrates the number of positive samples predicted to be negative and TN (True Negative) represents the number of negative samples predicted to be negative.

**Data availability.** The dataset analysed during the current study is available in the Github repository, https://github.com/YangzlTHU/C-EMRs.

### Acknowledgements

This research is supported by the National Natural Science Foundation of China (No. U1536201, No. U1536207 and No. U1636113).

### Author Contributions

Project design and implementation were conceived by Z.Y., Y.J., Y.S. and Y.H. Data collection was performed by Z.Y. and P.L. Data analysis was performed by Z.Y., Y.H. and Y.Z. The experiments and programming were performed by Z.Y. and Y.J. Manuscript drafting and editing was performed by Y.S. and Z.Y. revised the manuscript. All authors reviewed the manuscript in its final form.

### Additional Information

**Competing Interests:** The authors declare no competing interests.

**Publisher's note:** Springer Nature remains neutral with regard to jurisdictional claims in published maps and institutional affiliations.